\PassOptionsToPackage{table}{xcolor}
\documentclass[11pt]{article}
\newcommand{\vftclaims}{\textsc{VFT-Claims}}
\newcommand{\vftverify}{\textsc{VFT-Verify}}
\newcommand{\vftagent}{\textsc{VFT-Agent}}

\usepackage[preprint]{acl}

\usepackage{times}
\usepackage{latexsym}

\usepackage[T1]{fontenc}
\usepackage[utf8]{inputenc}
\usepackage{microtype}
\usepackage{inconsolata}
\usepackage{textcomp}

\usepackage{booktabs}
\usepackage{tabularx}
\usepackage{amsmath}
\usepackage{pifont}

\usepackage{graphicx}

\usepackage{doi}
\usepackage{url}
\usepackage{xurl}

\usepackage[table]{xcolor}
\usepackage[most]{tcolorbox}
\usepackage{fvextra}

\definecolor{SystemPromptColor}{HTML}{8DA0CB} 
\definecolor{UserPromptColor}{HTML}{FDBE85}   

\definecolor{TerminalPromptColor}{HTML}{4C78A8}

\newcommand{\cmark}{\textcolor{green!60!black}{\ding{51}}}    
\newcommand{\xmark}{\textcolor{red!80!black}{\ding{55}}}      
\newcommand{\pmark}{\textcolor{orange!70!black}{\boldmath$\triangle$}}

\title{Do Companies Deliver on Their IPO Claims?}

\definecolor{MarketColor}{HTML}{1f77b4}
\definecolor{ProductColor}{HTML}{ff7f0e}
\definecolor{FinancialColor}{HTML}{2ca02c}
\definecolor{OrgColor}{HTML}{d62728}
\definecolor{SocietalColor}{HTML}{9467bd}
\newcolumntype{C}[1]{>{\columncolor{#1}}p{3pt}}

\usepackage{marvosym}
\author{
\textbf{Vidhyakshaya Kannan\textsuperscript{1*\Letter} \quad Shiv Banafar\textsuperscript{2*} \quad Michael Galarnyk\textsuperscript{2*}} \\
\textbf{Sagnik Nandi\textsuperscript{2} \quad Soumeya Kannan\textsuperscript{2} \quad Shauray Arora\textsuperscript{2}} \\
\textbf{Abanoub Abdelmalak\textsuperscript{2} \quad Sudheer Chava\textsuperscript{2}} \\[0.75em]
\textsuperscript{1}Stanford University \qquad \textsuperscript{2}Georgia Institute of Technology \\
\Letter\ Corresponding Author: \texttt{kannanv@stanford.edu} \\
\textsuperscript{*}Indicates equal contribution
}

\begin{document}
\maketitle
\raggedbottom

\begin{abstract}
Initial public offering (IPO) filings describe what a company does and
what it intends to become. Yet whether individual forward-looking claims
are subsequently realized is difficult to assess. We introduce Vision
Follow-Through (VFT), a benchmark and agentic pipeline that separates claim extraction,
category assignment, verifiability, and evidence-grounded follow-through
assessment. \vftclaims{} contains 5,066 human-annotated claim spans from
50 IPO filings, while \vftverify{} adds verifiability tiers,
follow-through outcomes, and cited later evidence for 1,000 claims.
Across language models, classifying human-annotated claims is substantially
easier than extracting claims from full filings. Extraction coverage also
varies sharply by category, with Organizational/Culture claims recovered
far less often than Market Opportunity claims. Follow-through assessment
poses a separate challenge: model judgments diverge from human assessments
in different ways rather than exhibiting a single consistent bias. These results show that reliable assessment of corporate
follow-through requires both finding the relevant claims and grounding
subsequent judgments in later evidence.
\end{abstract}

\section{Introduction}

\begin{figure}[t]
    \centering
    \includegraphics[width=\columnwidth]{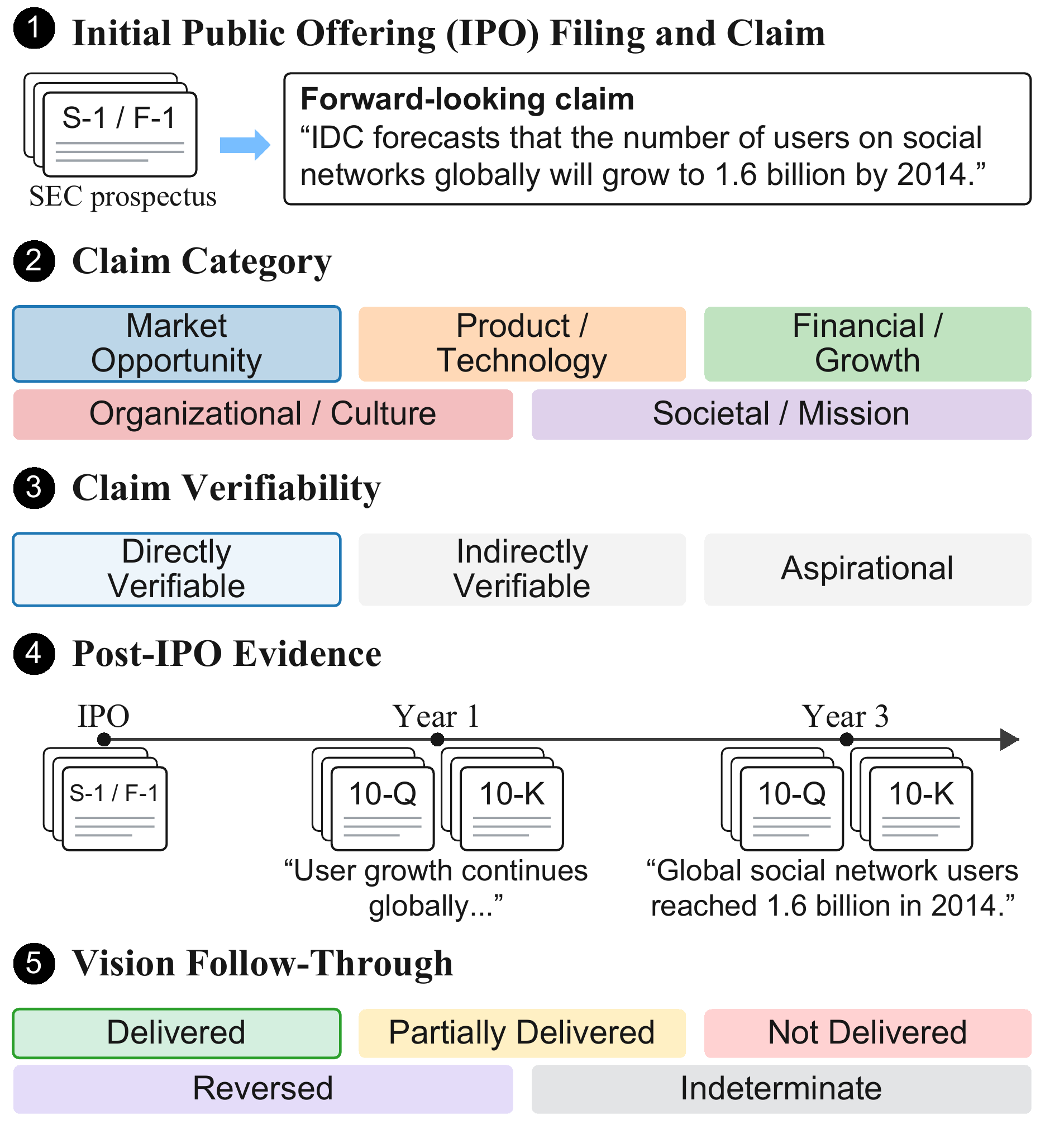}
    \caption{Vision Follow-Through (VFT) separates three research questions: which IPO statements are claims, what they concern, and whether later evidence establishes follow-through.}
    \label{fig:VFT}
\end{figure}

Initial public offering (IPO) filings describe what a company does and what
it intends to become. The language in these filings has been linked to
returns, volatility, underpricing, offer-price revisions, withdrawal, and
longer-term growth and innovation
\citep{hanley2010information,loughran2013ipo,sharpe-decker-2022-prospectus,WUORINEN2025e00526}.
These relationships, however, do not tell us whether individual
forward-looking claims are subsequently realized.

Assessing those claims in long, multimodal IPO filings
\citep{10.1145/3770855.3817580} is difficult because they can vary in
specificity and verifiability, and their scope may span several sentences.
Relevant evidence may appear years later across long filings. Missing
evidence must also be distinguished from evidence of non-delivery, while
some claims remain unresolved because the relevant period has not yet
elapsed. A useful benchmark must therefore evaluate both claim extraction
from the filing and whether later evidence supports follow-through.

\begin{table*}[t]
\centering
\footnotesize
\setlength{\tabcolsep}{2.5pt}
\renewcommand{\arraystretch}{1.12}

\begin{tabularx}{\textwidth}{@{}
>{\raggedright\arraybackslash}p{4.35cm}
>{\raggedright\arraybackslash}p{1.85cm}
>{\raggedright\arraybackslash}p{2.14cm}
>{\raggedright\arraybackslash}p{1.6cm}
@{\hspace{3pt}}c
@{\hspace{3pt}}c
@{\hspace{3pt}}c
@{\hspace{3pt}}c
@{\hspace{3pt}}c@{}}
\toprule
\textbf{Dataset / System} &
\shortstack{\textbf{Claim} \textbf{Source}} &
\shortstack{\textbf{Claim} \textbf{Focus}} &
\textbf{Evidence} &
\textbf{Issuer} &
\textbf{Future} &
\textbf{Human} &
\textbf{Agent} &
\textbf{Temporal} \\
\midrule

\textbf{NumClaim} \citep{10.1145/3340531.3412100} &
Reports & Numerical & -- & \xmark & \pmark & \cmark & \xmark & \xmark \\

\textbf{FinNumClaim} \citep{shah-etal-2024-numerical} &
Reports/Calls & Numerical & -- & \pmark & \pmark & \cmark & \xmark & \xmark \\

\textbf{FinDVer} \citep{zhao-etal-2024-findver} &
10-K/10-Q & Financial & Text/Tables & \cmark & \xmark & \cmark & \xmark & \xmark \\

\textbf{AUDITFLOW} \citep{wang2026auditflowexecutablesymbolicenvironments} &
XBRL & Consistency & XBRL/Rules & \cmark & \xmark & \cmark & \cmark & \xmark \\

\textbf{ML-Promise} \citep{seki-etal-2025-ml} &
ESG reports & Promises & Text/Images & \cmark & \cmark & \cmark & \xmark & \xmark \\

\textbf{TSVer} \citep{strong-vlachos-2025-tsver} &
Articles & Factual & Time-series & \xmark & \xmark & \cmark & \xmark & \pmark \\

\midrule
\textbf{VFT} \textbf{(Ours)} &
IPO firms & Forward-looking & SEC/Web & \cmark & \cmark & \cmark & \cmark & \cmark \\

\bottomrule
\end{tabularx}

\caption{
Comparison of claim detection and verification datasets/systems.
\textit{Issuer}: claims by the evaluated firm.
\textit{Future}: claims about future actions, goals, performance, or commitments.
\textit{Human}: human-labeled claims or outcomes.
\textit{Agent}: agents gather and assess evidence against human or benchmark labels.
\textit{Temporal}: later evidence determines what occurred after the claim.}
\label{tab:paper_comparison}
\end{table*}

We introduce Vision Follow-Through (VFT), a benchmark for studying
claim-level follow-through in IPO filings. Figure~\ref{fig:VFT} summarizes
the task: VFT follows claims from the IPO filing into subsequent evidence,
covering claim extraction, category assignment, verifiability, and
follow-through outcome. Human annotators provide reference annotations for
these stages; we then evaluate models against them. \vftclaims{} contains
5,066 human-annotated forward-looking claim spans from 50 IPO filings,
while \vftverify{} contains 1,000 claims with human verifiability tiers and
follow-through outcomes using later evidence.

Model performance differs substantially across these stages. Models are
better at assigning categories to human-annotated claims than at extracting
those claims from full IPO filings, and extraction coverage is particularly
weak for Organizational/Culture claims. Follow-through presents a separate
challenge: model judgments can differ substantially from human assessments,
and these disagreements take different forms across models rather than
reflecting a single consistent bias. These results show that performance on
supplied claims can mask failures in both claim extraction and
evidence-grounded follow-through assessment.

Our contributions are: (1) a temporal, cross-document benchmark
that separates claim extraction, category assignment, verifiability, and
evidence-grounded follow-through assessment; (2) two human-annotated
datasets, \vftclaims{} for forward-looking claim extraction and categorization
and \vftverify{} for verifiability and follow-through using evidence;
and (3) an evaluation of LLMs across claim extraction, category
assignment, and evidence-grounded follow-through verification.

\section{Related Work}
\label{sec:related-work}

\begin{table*}[t]
\centering
\small
\setlength{\tabcolsep}{4pt}
\renewcommand{\arraystretch}{1.05}

\begin{tabularx}{\textwidth}{@{}>{\raggedright\arraybackslash}p{3.2cm} X X@{}}
\toprule
\textbf{Category} & \textbf{Definition} & \textbf{Example Claim} \\
\midrule

\textbf{Market Opportunity}\newline
\textbf{(MO)} &
Total addressable market (TAM), market expansion, competitive positioning, or long-term industry opportunity. &
\textit{We believe we have a significant opportunity to better monetize our games in international markets.} \\

\textbf{Product/Technology}\newline
\textbf{(PT)} &
Proprietary technologies, product capabilities, platform architecture, or technical differentiation. &
\textit{We regularly update our games to add new content and improve monetization.} \\

\textbf{Financial/Growth}\newline
\textbf{(FG)} &
Profitability pathways, growth metrics, scalability, revenue expansion, or financial targets. &
\textit{We expect to achieve sustained revenue growth through increased user engagement.} \\

\textbf{Organizational/Culture}\newline
\textbf{(OC)} &
Governance reforms, leadership changes, compliance posture, cultural transformation, or regulatory collaboration. &
\textit{We continue to believe that making equity awards a key component of executive compensation aligns the executive team with the long-term interests of our stockholders.} \\

\textbf{Societal/Mission}\newline
\textbf{(SM)} &
Environmental impact, worker empowerment, public benefit, or mission framing. &
\textit{Our mission is to connect the world through games.} \\

\bottomrule
\end{tabularx}

\caption{Claim Categories}
\label{tab:claim_categories}
\end{table*}

Table~\ref{tab:paper_comparison} contrasts VFT with claim-detection and verification benchmarks, none of which follow forward-looking claims into later filings.

\subsection{Financial NLP and Benchmarking}
\label{subsec:financial-nlp}
Financial NLP includes specialized lexicons
\citep{loughran2011liability} and models such as FinBERT
\citep{araci2019finbert}, FLANG, BloombergGPT, and FinGPT
\citep{shah2022fluemeetsflangbenchmarks,wu2023bloomberggpt,yang2023fingpt}.
Benchmarks span sentiment, question answering, classification, and inference
\citep{FiQA,shah2025wordsuniteworldunified,magomere-etal-2025-finnli,
wang-etal-2025-surf}. Financial language has been studied for
forward-looking information in filings \citep{li2010forward}, earnings-call signals \citep{choi2026textalphallmstrack}, and whether corporate tech talk predicts later innovation \citep{chava2025walkingtechtalk}; forecasting benchmarks span earnings
surprises from transcripts \citep{koval-etal-2023-forecasting} and LLM
agents on prediction-markets \citep{ye2026forecastdojoreplayableenvironmentsbenchmarking}.

\subsection{Claim Extraction and Verification}
\label{subsec:claim-verification}

Verification spans Wikipedia \citep{thorne2018fever}, scientific
claims \citep{wadden2020scifact}, financial text and tables
\citep{zhao-etal-2024-findver}, and executable numerical checks
\citep{wang2026auditflowexecutablesymbolicenvironments}. Claim-level
methods identify check-worthy statements \citep{hassan2017claimbuster},
evaluate atomic facts
\citep{min2023factscorefinegrainedatomicevaluation}, check generated
claims against source documents \citep{tang-etal-2024-minicheck}, and
decompose verification into sub-tasks
\citep{pan2023factcheckingcomplexclaimsprogramguided}. These methods
assume static evidence; temporal verification instead reasons over post-claim
evidence \citep{strong-vlachos-2025-tsver,allein-etal-2023-implicit}.

\section{Annotation Framework}
\label{sec:taxonomy}

\begin{table*}[t]
\centering
\small
\setlength{\tabcolsep}{4pt}
\renewcommand{\arraystretch}{1.05}

\begin{tabularx}{\textwidth}{@{}>{\raggedright\arraybackslash}p{2.55cm} X X@{}}
\toprule
\textbf{Tier} & \textbf{Definition} & \textbf{Example Claim} \\
\midrule

Directly Verifiable &
Contains a measurable or externally observable outcome. &
``We will launch product X in COUNTRY by Q4 YEAR.'' \\

Indirectly Verifiable &
Expresses strategic intent that can be assessed from later evidence, but lacks a precise benchmark. &
``We aim to expand into international markets.'' \\

Aspirational &
Expresses mission, value, or ambition language without a clear external verification condition. &
``We strive to revolutionize sustainable living.'' \\

\bottomrule
\end{tabularx}
\caption{Verifiability tiers for forward-looking IPO claims.}
\label{tab:verifiability_tiers}
\end{table*}

\begin{table}[t]
\centering
\small
\setlength{\tabcolsep}{4pt}
\renewcommand{\arraystretch}{1.05}

\begin{tabularx}{\columnwidth}{@{}p{1.6cm}X@{}}
\toprule
\textbf{Label} & \textbf{Decision rule} \\
\midrule

Delivered &
Evidence establishes realization consistent with the claim's
scope and applicable deadline. \\

Partially Delivered &
Evidence establishes some, but not all, of the claimed realization
by the covered date. \\

Not Delivered &
Sufficient evidence establishes non-realization by the
applicable deadline or assessment date. \\

Reversed &
Evidence establishes that the firm abandoned or contradicted
the claim. \\

Indeterminate &
Evidence is insufficient, the relevant period is ongoing, or the claim
is not assessable. \\

\bottomrule
\end{tabularx}
\caption{Follow-through outcomes.}
\label{tab:outcomes}
\end{table}

To assess whether firms follow through on IPO claims, we use three label types: claim categories, verifiability tiers, and follow-through outcomes. Categories expose where models under-extract claim types; outcomes distinguish partial progress, non-delivery, reversal, and unresolved evidence.

\subsection{Claim Categories}

Claims are assigned categories derived from recurring themes in filings
(Table~\ref{tab:claim_categories}). \textbf{Market Opportunity (MO)} captures market positioning and expansion
\citep{loughran2013ipo};
\textbf{Product/Technology (PT)} covers products, technology, and innovation
\citep{WUORINEN2025e00526};
\textbf{Financial/Growth (FG)} covers financial and operating growth
\citep{li2010forward};
\textbf{Organizational/Culture (OC)} covers governance, leadership, workforce,
and culture \citep{merkl2007impression}; and
\textbf{Societal/Mission (SM)} covers social, environmental, and
mission claims \citep{christensen2021mandatory}.

\subsection{Verifiability Tiers}

Table~\ref{tab:verifiability_tiers} distinguishes \textit{Directly
Verifiable}, \textit{Indirectly Verifiable}, and \textit{Aspirational}
claims. Unlike work that identifies check-worthy factual statements
\citep{hassan2017claimbuster} or assigns supported, refuted, and
insufficient-evidence judgments \citep{thorne2018fever}, these tiers
describe how directly a forward-looking claim can be assessed from later
evidence. \textit{Directly Verifiable} claims can be checked against
specific events or metrics; \textit{Indirectly Verifiable} claims require
inference and evidence synthesis
\citep{pan2023factcheckingcomplexclaimsprogramguided}; and
\textit{Aspirational} claims lack a clear external verification condition.

\subsection{Follow-Through Outcomes}
\label{sec:outcome-policy}

Table~\ref{tab:outcomes} defines the follow-through outcomes. We assess each claim against its wording, deadline, and the dates covered by available evidence. Missing evidence alone does not establish non-delivery, and a claim whose deadline has not passed cannot be labeled \textit{Not Delivered}. Mentioning the relevant topic without resolving the claim is insufficient; \textit{Reversed} requires explicit evidence of abandonment or contradiction. Verifiability and outcome remain distinct: a \textit{Directly Verifiable} claim may be \textit{Indeterminate}
when evidence is unavailable.

\section{Dataset Construction}
\label{sec:dataset}
\label{sec:annotation}
\label{sec:guidelines}

VFT uses 50 IPO filings (S-1/F-1) filed with the Securities and Exchange
Commission (SEC) from 2011--2025. \vftclaims{} contains 5,066
human-annotated forward-looking claim spans; \vftverify{} contains 1,000
of these claims with verifiability tiers, follow-through outcomes, and
cited evidence. Follow-through evidence comes from later 10-Ks, 10-Qs,
and amended IPO filings.

\paragraph{Claim annotation.}
\label{sec:pilot}
Each filing is independently annotated by three annotators, who identify
forward-looking claim spans and assign a category. We reconcile overlapping
spans using character-level intersection-over-union
\citep{deyoung-etal-2020-eraser} and retain spans supported by at least two
annotators.

\paragraph{Follow-through annotation.}
For each claim in \vftverify{}, annotators assign a verifiability tier and
follow-through outcome using later filings, with evidence sufficient to
reproduce the assignment \citep{schlichtkrull2023averitec}.

\paragraph{Dataset records.}
\vftclaims{} records each consensus claim span, category, and source
location. \vftverify{} adds the verifiability tier, follow-through outcome,
and cited evidence, separating broad claim coverage from detailed evidence
review \citep{deyoung-etal-2020-eraser,zuo2022beyondbelief}.
Dataset statistics are reported in Appendix~\ref{app:dataset-statistics}.

\paragraph{Annotation agreement.}
\label{sec:annotation-agreement}
We measure span agreement using exact match and token overlap, and categorical agreement. Among 4,675 spans aligned across all three annotators, 4,262 (91.2\%) receive unanimous category labels. We use
character-level overlap to construct consensus spans.

\section{The VFT Agent}
\label{sec:pipeline}

\begin{table*}[t]
\centering\small
\setlength{\tabcolsep}{5pt}
\begin{tabularx}{\textwidth}{Xrrrrr}
\toprule
& \multicolumn{3}{c}{\textbf{Claim Extraction}} &
\multicolumn{2}{c}{\textbf{Category Assignment}} \\
\cmidrule(lr){2-4}\cmidrule(lr){5-6}
\textbf{Model} & \textbf{P} & \textbf{R} & \textbf{F$_1$} &
\textbf{Acc.} & \textbf{Macro-F$_1$} \\
\midrule
Claude Haiku 4.5 & \textbf{17.48} & 65.61 & \textbf{27.60} & 72.36 & 70.69 \\
Gemma 4 31B & 16.39 & \textbf{82.91} & 27.37 & 75.37 & 73.72 \\
GPT-6 Luna & 15.27 & 72.94 & 25.25 & \textbf{76.55} & \textbf{74.81} \\
GPT-oss 20B & 10.51 & 30.14 & 15.59 & 64.23 & 62.89 \\
GPT-oss 120B & 9.30 & 59.93 & 16.09 & 67.53 & 64.46 \\
Llama 3.1 8B & 2.26 & 63.38 & 4.37 & 60.66 & 57.03 \\
Llama 3.1 70B & 10.39 & 53.95 & 17.42 & 71.87 & 71.89 \\
Nemotron 3 Nano 4B & 4.02 & 17.69 & 6.56 & 66.42 & 63.05 \\
Nemotron 3 Nano 30B A3B & 1.70 & 65.18 & 3.32 & 57.36 & 56.09 \\
Nemotron 3.5 Super VL & 10.07 & 80.79 & 17.90 & 73.06 & 73.34 \\
\bottomrule
\end{tabularx}

\caption{Model performance on \vftclaims{} across two tasks.
\textbf{Claim extraction} tests whether models can locate forward-looking
claim spans within full IPO filings, scored by location-constrained
token-overlap matching. \textbf{Category assignment} tests whether models
can label human-annotated spans with one of five categories, isolating
labeling from extraction errors. Bold marks the best score per column.}
\label{tab:model-results}
\end{table*}

\begin{figure}[!t]
    \centering
    \includegraphics[width=\columnwidth]{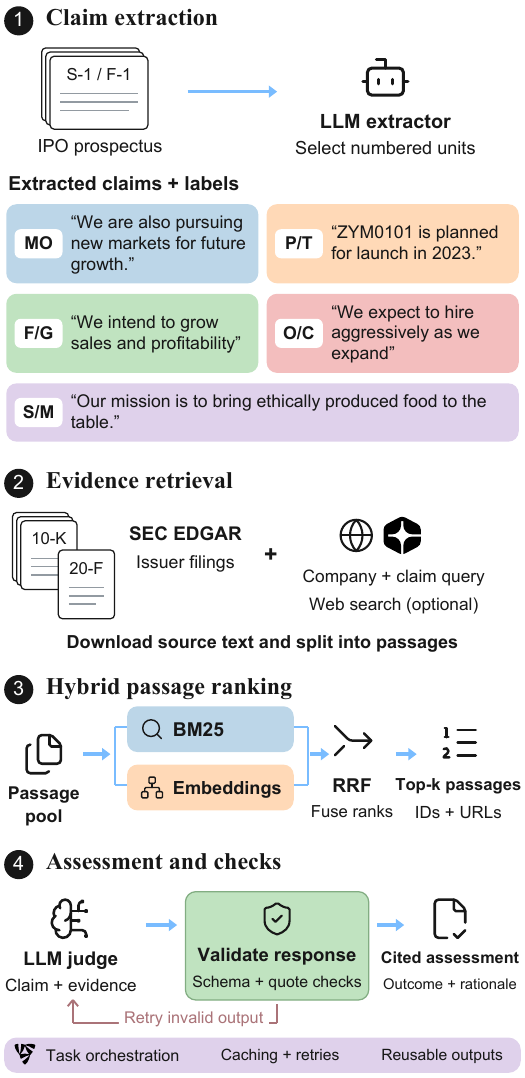}
    \caption{\vftagent{} extracts and labels claims, retrieves SEC and optional web evidence, ranks passages, and produces validated follow-through assessments.}
    \label{fig:vft-agent}
\end{figure}

\vftagent{} runs the benchmark through four stages: claim extraction,
labeling, evidence retrieval, and follow-through assessment
(Figure~\ref{fig:vft-agent}). Intermediate outputs are saved at each
stage, and supplying human-annotated claims bypasses extraction so missed
claims do not inflate later errors.

\subsection{Models}

We evaluate the following LLMs on \vftclaims{}: Claude Haiku 4.5
\citep{anthropic2025claude}, Gemma 4 31B
\citep{gemmateam2026gemma4technicalreport}, GPT-6 Luna
\citep{openai2026-sol-luna}, GPT-oss 20B, GPT-oss 120B
\citep{openai2025gptoss120bgptoss20bmodel}, Llama-3.1 8B, Llama-3.1 70B
\citep{grattafiori2024llama3herdmodels}, Nemotron 3 Nano 4B, Nemotron 3 Nano 30B A3B, and Nemotron 3.5 Super VL \citep{nvidia-nemotron3}.

\subsection{Claim Extraction and Labeling}
\label{sec:extraction}
\label{sec:classification}
The extractor reads IPO filing text as numbered, sentence-like units and
returns identifiers of claim-bearing units, which we reconstruct into
claim text and location. IPO-Mine \citep{10.1145/3770855.3817580} is
exposed as a tool for VFT-Agent: given a central index key, it retrieves the filing from SEC EDGAR and returns its text.
A separate labeling stage assigns each claim a category and verifiability
tier.

\subsection{Evidence Retrieval}
\label{sec:agent-retrieval}
For each claim, the retriever pools passages from the issuer's 10-K and
20-F filings and web pages retrieved
via You.com \citep{youcom_search}, a web search API for AI agents,
queried with the company name and claim text. Using the claim as query, passages are ranked by BM25 lexical
matching \citep{10.1561/1500000019} and sentence-transformer cosine
similarity \citep{reimers-gurevych-2019-sentence}, fused via reciprocal
rank fusion (RRF; \citealp{cormack2009rrf}). The verifier receives up to
eight top-ranked passages with source identifiers.

\begin{table*}[t]
\centering\small
\setlength{\tabcolsep}{3pt}
\begin{tabularx}{\textwidth}{Xrrrrrrrrrr}
\toprule
& \multicolumn{5}{c}{\textbf{Extraction recall}}
& \multicolumn{5}{c}{\textbf{Category F$_1$}} \\
\cmidrule(lr){2-6}\cmidrule(lr){7-11}
\textbf{Model}
& \textbf{FG} & \textbf{MO} & \textbf{PT} & \textbf{OC} & \textbf{SM}
& \textbf{FG} & \textbf{MO} & \textbf{PT} & \textbf{OC} & \textbf{SM} \\
\midrule
Claude Haiku 4.5
& 65.07 & 75.08 & 66.70 & 40.31 & 63.10
& 80.40 & 66.67 & 66.67 & 65.60 & 74.10 \\

Gemma 4 31B
& \textbf{80.05} & \textbf{92.26} & \textbf{86.70} & 58.13 & \textbf{88.28}
& 80.63 & \textbf{75.00} & 72.93 & 64.37 & 75.69 \\

GPT-6 Luna
& 71.90 & 80.42 & 76.54 & 45.21 & 78.62
& \textbf{84.33} & 71.59 & \textbf{73.84} & 63.84 & \textbf{80.45} \\

GPT-oss 20B
& 34.98 & 27.63 & 25.41 & 24.28 & 30.34
& 69.47 & 66.51 & 58.94 & 59.87 & 59.68 \\

GPT-oss 120B
& 67.01 & 64.01 & 51.03 & 42.76 & 45.17
& 71.87 & 70.49 & 70.46 & 55.03 & 54.47 \\

Llama 3.1 8B
& 63.60 & 64.40 & 62.59 & 60.36 & 64.48
& 71.01 & 51.47 & 59.07 & 51.03 & 52.55 \\

Llama 3.1 70B
& 51.33 & 62.77 & 57.51 & 26.73 & 64.48
& 77.04 & 69.29 & 66.17 & 71.52 & 75.40 \\

Nemotron 3 Nano 4B
& 18.82 & 16.33 & 19.68 & 13.59 & 15.52
& 73.15 & 62.16 & 64.39 & 54.89 & 60.64 \\

Nemotron 3 Nano 30B A3B
& 68.29 & 64.47 & 61.84 & \textbf{62.36} & 60.69
& 55.39 & 60.90 & 58.33 & 54.93 & 50.89 \\

Nemotron 3.5 Super VL
& 79.15 & 87.46 & 84.32 & 60.36 & 83.45
& 77.56 & 73.02 & 65.48 & \textbf{76.13} & 74.52 \\

\bottomrule
\end{tabularx}

\caption{Extraction recall compares model-extracted claims with human annotations; Category F$_1$ evaluates category assignment when models are given the human-annotated claims.}
\label{tab:category-results}
\end{table*}

\subsection{Follow-Through Assessment}
\label{sec:follow-through-assessment}

Given the claim and retrieved passages, the verifier returns an outcome,
confidence, rationale, and quotations linked to passage identifiers.
VFT-Agent validates the schema and checks each quotation against
its cited passage, retrying on invalid output. These checks validate output
structure and quotation provenance, not whether the cited evidence
semantically supports the judgment. Missing evidence and execution failures
are recorded separately from produced judgments. We orchestrate runs with
Flyte \citep{flyte}, a durable workflow runtime whose retries
and caching let the pipeline resume after API errors or timeouts without
restarting. Appendix~\ref{app:pipeline-details} details selection,
evidence windows, execution settings, and assessment rules.

\section{Results and Error Analysis}
\label{sec:results}

\begin{table*}[t]
\centering
\small
\renewcommand{\arraystretch}{0.9}
\setlength{\tabcolsep}{3.5pt}

\begin{tabular}{
l
r r r r r
!{\vrule width 0.7pt\hspace{3pt}}
r r r r r
}
\toprule

& \multicolumn{5}{c}{\textbf{Outcome Share (\%)}}
& \multicolumn{5}{c}{\textbf{Delivered by Category (\%)}} \\

\cmidrule(lr){2-6}
\cmidrule(lr){7-11}

\textbf{Model} &
\textbf{Delivered} &
\shortstack{\textbf{Partially}\\\textbf{Delivered}} &
\shortstack{\textbf{Not}\\\textbf{Delivered}} &
\textbf{Reversed} &
\textbf{Indeterminate} &
\textbf{FG} & \textbf{MO} & \textbf{PT} & \textbf{OC} & \textbf{SM} \\

\midrule

\textit{Human}
& 36.9 & 29.6 & 3.1 & 0.7 & 29.7 & 47.1 & 25.8 & 33.1 & 27.4 & 39.7 \\

\midrule
Claude Haiku 4.5     & 37.5 & 10.9 & 0.7 & 0.4 & 17.5 & 45.7 & 33.5 & 31.9 & 30.8 & 33.3 \\ Gemma 4 31B          & 50.7 & 7.7 & 1.5 & 0.9 & 25.3 & 54.0 & 41.2 & 48.8 & 46.2 & 64.3 \\
GPT-6 Luna           & 20.8 & 24.8 & 0.2 & 0.2 & 40.4 & 24.2 & 16.3 & 20.6 & 18.8 & 20.6 \\
GPT-oss 20B          & 28.0 & 5.2 & 3.0 & 0.3 & 47.1 & 31.1 & 17.6 & 31.9 & 27.4 & 32.5 \\
GPT-oss 120B         & 35.9 & 5.9 & 2.4 & 0.4 & 39.7 & 36.2 & 26.7 & 38.8 & 40.2 & 43.7 \\
Llama 3.1 8B         & 4.5 & 46.6 & 8.9 & 0.0 & 2.0 & 1.9 & 4.1 & 4.4 & 5.1 & 12.7 \\
Llama 3.1 70B          & 46.5 & 16.7 & 1.2 & 0.3 & 19.2 & 46.0 & 34.4 & 51.2 & 47.9 & 61.9 \\
Nemotron 3.5 Super VL & 37.1 & 13.7 & 0.7 & 0.2 & 34.0 & 39.4 & 27.1 & 41.2 & 32.5 & 46.8 \\
\bottomrule
\end{tabular}

\caption{
Follow-through verification on the 1,000 \vftverify{} claims with SEC and web evidence.
\textbf{Outcome Share (\%)}: share of all claims assigned to each outcome.
\textbf{Delivered by Category (\%)}: share of claims within each category labeled Delivered.
Appendix~\ref{app:verification-coverage} reports coverage and scoring details.
}
\label{tab:verification-results}
\end{table*}

VFT exposes errors at three stages: claim extraction, category assignment, and follow-through assessment. We evaluate each stage separately so that downstream performance is not conflated with upstream claim-selection errors.

\subsection{Evaluation Setup}
\label{sec:results-setup}

We evaluate claim extraction and category assignment on \vftclaims{}.
Extraction is scored by location-constrained token overlap with
human-annotated claims; category assignment labels those claims directly,
isolating labeling errors. Follow-through is evaluated on the 1,000
\vftverify{} claims. Annotators cited later SEC filings
(Table~\ref{tab:evidence-coverage}); each claim's SEC and You.com passages
were retrieved once and shared by all models.

\subsection{Extraction and Category Assignment}
\label{sec:results-extraction}
\label{sec:results-category}

Table~\ref{tab:model-results} shows a gap between finding claims in
full filings and categorizing supplied claims. Extraction precision remains
below 18\% for every model and F$_1$ never exceeds 27.60, while category
assignment is substantially stronger. No model leads both tasks: Claude
Haiku 4.5 has the highest extraction F$_1$, Gemma 4 31B the highest recall,
and GPT-6 Luna the highest category accuracy and Macro-F$_1$. Strong
performance on supplied claims therefore does not imply reliable extraction. Larger models do not consistently improve extraction performance. GPT-oss 120B and Nemotron 3 Nano 30B A3B increase
recall with little or no F$_1$ improvement, while Llama 3.1 70B improves
F$_1$ despite lower recall. Claim selection remains the primary bottleneck.

\paragraph{Which claims are missed?}
Table~\ref{tab:category-results} shows strong category effects.
Organizational/Culture has the lowest extraction recall for nine of ten
models; only Nemotron 3 Nano 30B A3B performs worse on Societal/Mission.
This pattern is weaker for category assignment, suggesting that these
claims are harder to identify in full filings than to label once supplied.

\subsection{Error Analysis}
\label{sec:results-analysis}

Extraction errors extend beyond missing claims. Models may
omit part of a claim, merge claims, combine discontinuous text,
or segment the source differently from human annotation. We
distinguish eight error types: inconsistent claim criteria, omitted
claims, incomplete claims, merged claims, discontinuous claims,
granularity mismatch, category misclassification, and truncated
generation. Appendix~\ref{app:diagnostic-examples} provides examples.

\subsection{Follow-Through Verification}
\label{sec:results-verification}

Table~\ref{tab:verification-results} compares model and human
follow-through assessments on 1,000 claims using identical SEC and You.com
evidence. Human assessments assign 36.9\% of claims as Delivered, 29.6\%
as Partially Delivered, and 29.7\% as Indeterminate. Model distributions
differ, with no consistent bias: some models assign Delivered far more often
than humans, while others favor Indeterminate or Partially Delivered. Most
models assign substantially fewer Partially Delivered outcomes than humans.
Similar aggregate outcome shares do not imply agreement on individual
claims. Full-cohort accuracy ranges from 18.1\% to 44.1\%, with Macro-F$_1$
from 12.1 to 36.2. Judgment coverage is above 83\% for most models, although
Claude Haiku 4.5 and Llama 3.1 8B reach only 67.0\% and 62.0\%, respectively,
due largely to invalid outputs. Appendix~\ref{app:verification-coverage}
reports coverage, scoring, and model settings.

\section{Conclusion}
\label{sec:conclusion}

VFT frames IPO follow-through as a claim-level evaluation problem with
5,066 human-annotated claims and 1,000 follow-through assessments. Results
show that category assignment on supplied claims is substantially easier
than extracting those claims from full filings, with Organizational/Culture
claims particularly difficult to recover. Follow-through assessment poses a
separate challenge: model judgments differ substantially from human
assessments, and these disagreements vary across models rather than
reflecting a single consistent bias. Reliable assessment therefore requires
both identifying the relevant claims and grounding later judgments in
evidence while distinguishing full delivery, partial progress, non-delivery, reversal, and uncertainty. VFT makes these failures measurable separately.

\section*{Limitations}
\label{sec:limitations}

\paragraph{Source and temporal coverage.}
Annual reports may omit evidence available in 8-Ks, proxy statements, or
sustainability reports, affecting Organizational/Culture and
Societal/Mission claims most. Filings are company-authored and not an
exhaustive record of realized behavior; missing evidence must remain
distinguishable from documented non-delivery. Observation horizons,
claim deadlines, and evidence cutoffs constrain every verdict.

\paragraph{Selection and generalization.}
The 50 IPO filings were selected for the presence of CEO or founder letters,
which contain a high volume of aspirational language. The 1,000 verification
claims were subsequently drawn from these filings. This selection may
overrepresent aspirational claims and firms that include such letters.
Human outcome shares therefore describe the selected claims, not a
population estimate of how often companies fulfill claims. Aggregate
shares can also agree while individual verdicts differ, so we report
claim-level agreement alongside distributional differences.

\paragraph{Scope and interpretation.}
Follow-through assessments reflect the available evidence
within a defined observation period and may not capture
all subsequent corporate actions. These assessments
should not be interpreted as comprehensive measures of
corporate credibility or as evidence of predictive
investment value.

\section*{Ethical Considerations}
VFT is designed to support transparent, evidence-grounded analysis of
corporate follow-through. Assessments retain cited evidence and distinguish
delivery, partial delivery, non-delivery, reversal, and unresolved cases so
that judgments remain inspectable rather than opaque.

\section*{Acknowledgments}
\label{sec:acknowledgments}

We thank Arnav Hiray for helpful comments. We thank Dell Technologies for
providing a Dell Pro Max T2 workstation equipped with an NVIDIA RTX PRO 6000
Blackwell GPU. We also thank Georgia Tech's Partnership for an Advanced
Computing Environment (PACE) for computational resources and HumanSignal for
providing Label Studio.





%
 
\bibliography{custom}

\appendix
\section{Dataset and Annotation Details}
\label{sec:annotation-guide}

\subsection{Annotation Interface}
\label{app:interface}
\begin{figure*}[t]
\centering
\includegraphics[
    width=\textwidth,
    trim=0 12bp 0 0,
    clip
]{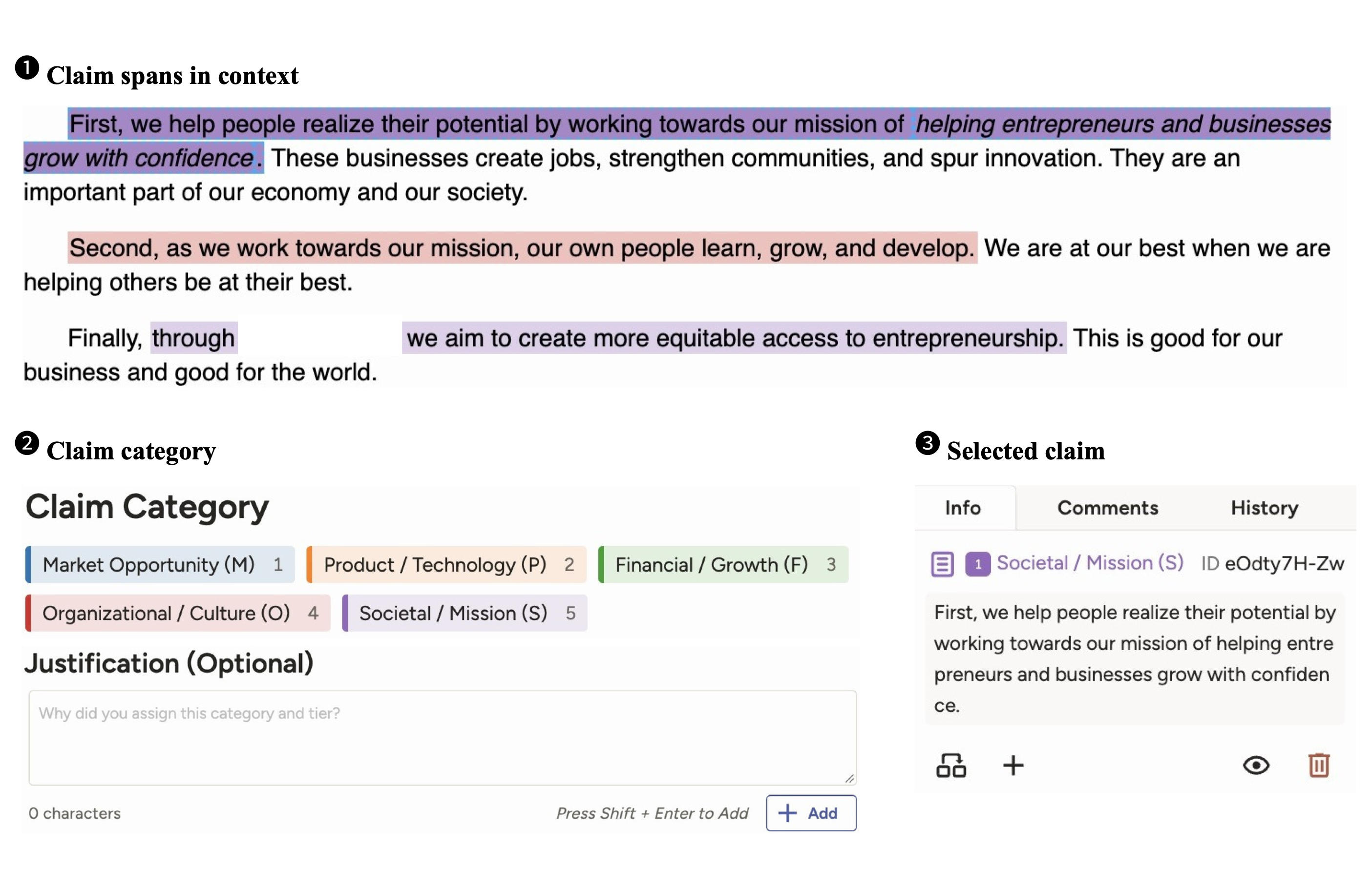}
\caption{VFT annotation interface in Label Studio. (1) Claim spans highlighted
in filing context. (2) Five claim categories and an optional justification
field. (3) The selected region's category and exact claim text.}
\label{fig:heartex-interface}
\end{figure*}

Figure~\ref{fig:heartex-interface} shows the Label Studio annotation interface used by annotators \cite{LabelStudio}.

\subsection{Dataset Statistics}
\label{app:dataset-statistics}
\begin{figure*}[t]
\centering
\includegraphics[width=\textwidth]{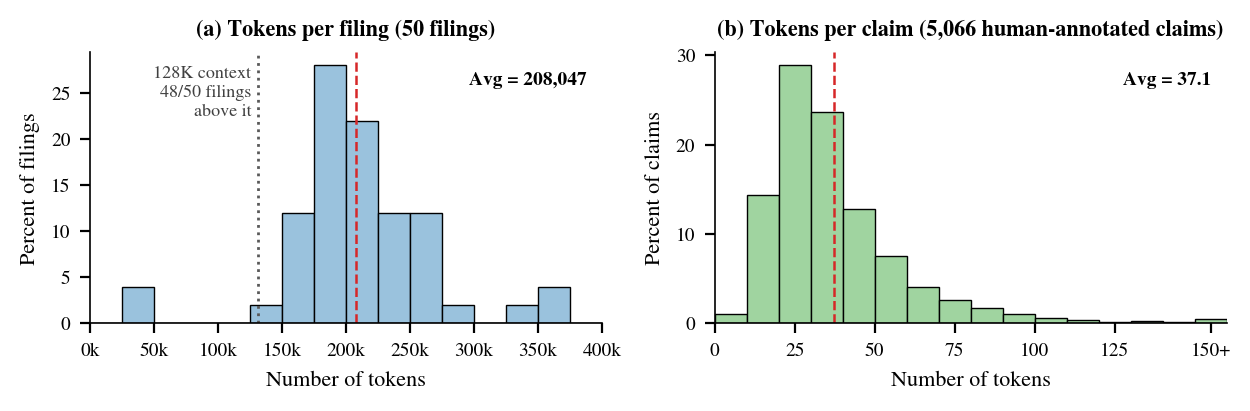}
\caption{Llama 3 token counts per filing (a) and per human-annotated claim (b).
Red dashed lines mark means; the dotted line marks a 128K-token context window.}
\label{fig:token-distribution}
\end{figure*}

This appendix summarizes \vftclaims{} and \vftverify{}
(Tables~\ref{tab:categories}--\ref{tab:evidence-coverage}).
For \vftclaims{}, we report category frequencies and document and claim
lengths. Figure~\ref{fig:token-distribution} highlights the difference in
scale: 48 of 50 filings exceed 128K tokens, while human-annotated claims
average 37 tokens.

For \vftverify{}, we report cited evidence sources and human-annotated
outcome counts. These outcome annotations are used in the evaluation
in Section~\ref{sec:results-verification}.

\begin{table*}[t]
\centering\small
\begin{tabular*}{\textwidth}{@{\extracolsep{\fill}}lrrrr}
\toprule
\textbf{Category} & \textbf{\vftclaims{}} & \textbf{Share} & \textbf{\vftverify{}} & \textbf{Share} \\
\midrule
\colorbox{FinancialColor!30}{\textbf{Financial/Growth}} & 2,110 & 41.7\% & 376 & 37.6\% \\
\addlinespace[3pt]
\colorbox{MarketColor!30}{\textbf{Market Opportunity}} & 1,292 & 25.5\% & 221 & 22.1\% \\
\addlinespace[3pt]
\colorbox{ProductColor!30}{\textbf{Product/Technology}} & 925 & 18.3\% & 160 & 16.0\% \\
\addlinespace[3pt]
\colorbox{OrgColor!30}{\textbf{Organizational/Culture}} & 449 & 8.9\% & 117 & 11.7\% \\
\addlinespace[3pt]
\colorbox{SocietalColor!30}{\textbf{Societal/Mission}} & 290 & 5.7\% & 126 & 12.6\% \\
\midrule
\textbf{Total categorized claims} & \textbf{5,066} & \textbf{100.0\%} & \textbf{1,000} & \textbf{100.0\%} \\
\bottomrule
\end{tabular*}
\caption{Category distribution.}
\label{tab:categories}
\end{table*}

\begin{table*}[t]
\centering\small
\begin{tabularx}{\textwidth}{Xrrr}
\toprule
\textbf{Measure} & \textbf{Mean} & \textbf{Median} & \textbf{Range} \\
\midrule
Llama 3 tokens per filing & 208,047.0 & 206,825.0 & 25,274 to 365,703 \\
Retained claims per filing & 101.3 & 86.5 & 8 to 379 \\
Llama 3 tokens per retained claim & 37.1 & 32.0 & 5 to 334 \\
\bottomrule
\end{tabularx}
\caption{Tokens are counted with the Llama 3 tokenizer (128K vocabulary) on cleaned HTML, including tables and financial notes.}
\label{tab:dataset-lengths}
\end{table*}

\begin{table}[t]
\centering\small
\begin{tabularx}{\columnwidth}{Xrr}
\toprule
\textbf{Cited source type} & \textbf{URLs} & \textbf{Claims} \\
\midrule
Annual reports (73 10-Ks; 5 20-Fs) & 78 & 834 \\
Quarterly reports (10-Q) & 8 & 75 \\
Registration statements (S-1, S-1/A, S-8) & 6 & 79 \\
Current reports and exhibits (8-K, 6-K) & 3 & 11 \\
Proxy statement (DEF~14A) & 1 & 7 \\
Annual report to shareholders (ARS) & 1 & 2 \\
\midrule
\textbf{Total unique URLs / claims} & \textbf{97} & \textbf{1,000} \\
\bottomrule
\end{tabularx}
\caption{Evidence coverage in \vftverify{}. A claim can cite multiple
source types, so the claim counts overlap. Coverage records citations;
it does not imply that every cited passage is sufficient to determine
follow-through.}
\label{tab:evidence-coverage}
\end{table}

\subsection{Span Alignment and Human-Annotated Boundaries}

We retain claim spans supported by at
least two of three annotators, yielding 5,066 human-annotated claims.
We align annotations within each filing using character-level
intersection-over-union (IoU $\geq 0.5$).

We group spans with complete-link agglomerative clustering, constrained to
at most one span per annotator. Two clusters can merge only if all span
pairs between them reach IoU $\geq 0.5$. At each step, we merge the eligible
pair whose weakest overlap is strongest (highest minimum pairwise IoU).
Ties are broken by ordered span keys: offsets, annotator identifier,
annotation identifier, and result identifier.

Each retained cluster is represented by the member with the highest summed
IoU against the other members, using the same tie-break rule. The
construction manifest preserves the contributing annotations, source
offsets, category votes, and representative span. Span support and category
assignment are separate decisions: disagreement about a claim's category
does not remove an otherwise supported span. The dataset contains 5,066 location-specific, human-annotated claim spans. Repeated wording is retained when it occurs at different locations. After normalizing case, punctuation, and whitespace, these spans correspond to 4,029 unique text strings across the corpus.
\subsection{Category Consensus and Annotation Diagnostic}
We assign categories by majority vote among the three annotators. Across
4,675 spans aligned among all three at IoU $\geq 0.5$, 4,262 category assignments are unanimous and 413 (8.83\%) are not: 406 have a two-to-one
split and seven have three different labels. These counts measure category
agreement on aligned spans before the category tie-break; they exclude
omissions and boundary disagreements and do not replace the reliability
analysis in Section~\ref{sec:annotation-agreement}.

Across the retained set, 459 claims receive votes for more than one
category. We grouped these disputed claims into four patterns using keyword
rules (Table~\ref{tab:disagreement-patterns}); a claim can show more than
one. \emph{Nested claims} contain multiple claims in one sentence that may
fit different categories, as in one issuer's to grow
its member base ``by growing the number of [...] locations, continuing to scale
our existing membership brands and launching and growing new membership
brands'' (Financial/Growth 2, Market Opportunity 2).
\emph{Ambiguous wording} uses terms associated with each category chosen.
\emph{Customer or geographic expansion} can be read as market opportunity
or as growth. \emph{Acquisitions} have no dedicated category, so annotators
labeled them by their stated purpose, often both growth and market entry.
Together, these patterns cover 79.7\% of disputed claims. Failure examples
in Appendix~\ref{app:diagnostic-examples} show a related confusion between
business objectives and the mechanisms used to reach them.

\begin{table}[t]
\centering\small
\begin{tabularx}{\columnwidth}{Xr}
\toprule
\textbf{Pattern} & \textbf{\% of disputed claims} \\
\midrule
Nested claims & 52.7\% \\
Ambiguous wording & 45.3\% \\
Customer or geographic expansion & 36.2\% \\
Acquisitions & 7.8\% \\
None of the above & 20.3\% \\
\bottomrule
\end{tabularx}
\caption{Patterns in disputed category labels, as a share of the 459
retained claims with votes for more than one category (all contributing
annotators). A claim can show several patterns, so shares sum to more than
100\%.}
\label{tab:disagreement-patterns}
\end{table}

\subsection{Verification Label Distributions}
\label{sec:verification-distributions}
Tables~\ref{tab:verify-tiers} and~\ref{tab:verify-outcomes} report the
verifiability tiers and human-annotated follow-through outcomes for the 1,000
completed assessments in \vftverify{}. Table~\ref{tab:verify-outcomes}
uses the same human-annotated outcome counts as Table~\ref{tab:verification-results};
these are reference labels, not model predictions.

\begin{table}[t]
\centering\small
\begin{tabularx}{\columnwidth}{Xrr}
\toprule
\textbf{Verifiability tier} & \textbf{Claims} & \textbf{Share} \\
\midrule
Directly Verifiable (DV) & 282 & 28.2\% \\
Indirectly Verifiable (IV) & 571 & 57.1\% \\
Aspirational (A) & 147 & 14.7\% \\
\textbf{Total} & \textbf{1,000} & \textbf{100.0\%} \\
\bottomrule
\end{tabularx}
\caption{Verifiability labels in the completed \vftverify{} selection. A tier describes how a claim can be checked; it does not imply that suitable evidence was found.}
\label{tab:verify-tiers}
\end{table}

\begin{table}[t]
\centering\small
\begin{tabular}{lrr}
\toprule
\textbf{Outcome} & \textbf{Claims} & \textbf{Share} \\
\midrule
Delivered & 369 & 36.9\% \\
Partially Delivered & 296 & 29.6\% \\
Not Delivered & 31 & 3.1\% \\
Reversed & 7 & 0.7\% \\
\midrule
Indeterminate & 297 & 29.7\% \\
\midrule
\textbf{Total} & \textbf{1,000} & \textbf{100.0\%} \\
\bottomrule
\end{tabular}
\caption{Human-annotated outcomes for the 1,000 \vftverify{} claims evaluated in
Section~\ref{sec:results-verification}.
\label{tab:verify-outcomes}}
\end{table}

\subsection{Evidence and Horizon Decision Rules}
Use Delivered only when evidence establishes the claimed realization;
use Partially Delivered when it establishes a proper part or attenuated
realization. Use Reversed for documented abandonment or contradiction of
the commitment. Not Delivered requires sufficient evidence of
non-realization by the applicable deadline or horizon. A missing passage,
generic discussion of the topic, or an unexpired deadline does not by
itself meet that requirement. Use Indeterminate when evidence or an
assessable condition is missing, and record the reason separately from
the claim's verifiability tier.

\begingroup
\definecolor{VFTLabelBlue}{HTML}{1F77B4}
\definecolor{VFTLabelOrange}{HTML}{FF7F0E}
\definecolor{VFTLabelGreen}{HTML}{2CA02C}
\definecolor{VFTLabelRed}{HTML}{D62728}
\definecolor{VFTLabelPurple}{HTML}{9467BD}
\definecolor{VFTInk}{HTML}{202B35}
\newenvironment{VFTPromptList}{\begin{list}{\textbullet}{%
  \setlength{\leftmargin}{14pt}\setlength{\labelsep}{5pt}%
  \setlength{\topsep}{3pt}\setlength{\partopsep}{0pt}%
  \setlength{\itemsep}{2pt}\setlength{\parsep}{0pt}}}{\end{list}}
\tcbset{vftprompt/.style={enhanced standard, breakable, arc=3pt, outer arc=3pt,
  boxrule=0.6pt, colback=white, coltext=VFTInk, coltitle=VFTInk,
  fonttitle=\small\bfseries, fontupper=\fontsize{9}{10.5}\selectfont,
  fontlower=\fontsize{9}{10.5}\selectfont,
  boxsep=0pt, left=7pt, right=7pt, top=7pt, bottom=7pt,
  toptitle=6pt, bottomtitle=6pt, titlerule=0pt,
  before skip=7pt, after skip=7pt, segmentation style={solid, opacity=0.3},
  before upper={\raggedright\setlength{\parindent}{0pt}\setlength{\parskip}{0pt}},
  before lower={\raggedright\setlength{\parindent}{0pt}\setlength{\parskip}{0pt}}
}}

\section{Prompts}\label{sec:prompts}\label{app:prompts}

\subsection{Claim Extraction}\label{fig:claim-tagging-prompt}

\begin{nolinenumbers}\begin{tcolorbox}[vftprompt, colframe=black, colback=white, colbacktitle=white, coltitle=black, title={Claim Extraction Prompt}, title after break={Claim Extraction Prompt (continued)}]
\textbf{\textsc{System message}}\par\smallskip
\noindent You are a meticulous securities-filing analyst annotating an IPO prospectus (SEC Form S-1 or F-1). The document is split into numbered units like \texttt{[U12]}. Your job is to TAG the units that express a CLAIM using the eligibility rules below and the supplied annotation guide. If supplementary guide wording conflicts with these eligibility rules, use these rules. Do not rewrite the text.\par
\par\smallskip
\par\noindent\textbf{Output rules (follow them strictly):}\par\nobreak
\begin{VFTPromptList}
\item For each claim, return ONLY the unit id(s) it spans and a category.
\item Do NOT echo the claim text, quotes, paraphrases, or explanations.
\item Group consecutive units that together form one claim into one entry. Every entry must correspond to one contiguous source passage. If multiple commitments occur within one indivisible unit, preserve the complete unit; do not invent sub-unit boundaries or duplicate an identical unit selection.
\item Use only unit ids that appear in the document.
\item Assign a \texttt{category} drawn from the taxonomy in the annotation guide.
\item Identify commitments and strategic claims about the company's market opportunity, products or technology, financial growth, organization or culture, and societal mission. Include plans, expectations, targets, and commitments to continuing practices. A present-tense statement must express a substantive position, commitment, or mission; exclude plain descriptions of facts or operations.
\item Governance reforms, leadership plans, compensation commitments, compliance commitments, and workforce or cultural commitments are eligible Organizational/Culture claims.
\item Tag only genuine claims; skip boilerplate, headings, definitions, offering terms, and safe-harbor language unless the guide says otherwise.
\item Also do not tag:
\begin{VFTPromptList}
\item financial statement lines, table rows, or accounting explanations (e.g. "Operating income (loss) (55,036)...", "The decrease was due to...");
\item routine procedural or legal descriptions that express no substantive commitment, strategic intent, or continuing organizational practice;
\item plain descriptions of how a product or process works that assert nothing about quality, position, growth, or direction.
\end{VFTPromptList}
\item If you are unsure whether a unit is a claim, do not tag it.
\end{VFTPromptList}
\par\smallskip
\noindent === ANNOTATION GUIDE ===\par
\noindent \texttt{\{guide\_text\}}\par
\noindent === END ANNOTATION GUIDE ===\par
\tcblower\textbf{\textsc{User message}}\par\smallskip
\noindent Tag the claim units in the following IPO prospectus (S-1 or F-1). Return only unit ids and categories.\par
\par\smallskip
\noindent \texttt{\{tagged\_document\}}\par
\par\smallskip
\noindent Respond with a single JSON object that matches the schema. Do not include any prose outside the JSON.\par
\end{tcolorbox}\end{nolinenumbers}

\subsection{Independent Label Assignment}\label{app:claim-label-prompts}

\begin{nolinenumbers}\begin{tcolorbox}[vftprompt, colframe=black, colback=white, colbacktitle=white, coltitle=black, title={Independent Label Assignment Prompt}, title after break={Independent Label Assignment Prompt (continued)}]
\textbf{\textsc{System message}}\par\smallskip
\noindent You assign labels to already-extracted IPO prospectus claims. Do NOT decide whether a span is a claim; assume every provided span is a claim. For each claim, assign exactly one subject category and exactly one verifiability tier from the annotation-guide taxonomy.\par
\par\smallskip
\par\noindent\textbf{Subject category:}\par\nobreak
\begin{VFTPromptList}
\item MO: Market Opportunity --- addressable markets, demand growth, market expansion, competitive positioning, and industry opportunity.
\item PT: Product / Technology --- products, platforms, proprietary technology, architecture, technical capabilities, differentiation, and innovation strategy.
\item FG: Financial / Growth --- profitability, revenue growth, scalability, margins, financial targets, and future operating performance.
\item OC: Organizational / Culture --- governance, leadership, compensation, compliance, workforce practices, and internal culture.
\item SM: Societal / Mission --- environmental impact, stakeholder value, public benefit, worker empowerment, and broader social or environmental mission.
\end{VFTPromptList}
\par\smallskip
\par\noindent\textbf{Verifiability tier:}\par\nobreak
\begin{VFTPromptList}
\item DV (Directly Verifiable): the claim specifies a measurable or externally observable outcome that can be checked against events or metrics in later filings.
\item IV (Indirectly Verifiable): the claim expresses strategic intent that can be assessed from later evidence but lacks a precise benchmark and requires inference.
\item A (Aspirational): the claim expresses a mission, value, or ambition without a clear external verification condition.
\end{VFTPromptList}
\par\smallskip
\noindent Assign the primary category that best describes the claim's subject. A numerical mention alone does not establish direct verifiability, and a directly verifiable claim can still have an Indeterminate outcome if evidence is unavailable. Set negated=true only for self-limiting claims that explicitly constrain expectations, such as not expecting profitability. This flag preserves the claim's meaning; it does not determine the outcome. Return a concise rationale tied to the text.\par
\tcblower\textbf{\textsc{User message}}\par\smallskip
\par\noindent\textbf{Assign category and verifiability labels to these already-extracted IPO claims:}\par\nobreak
\par\smallskip
\noindent \texttt{\{claim\_id\}} | \texttt{\{section\}}: \texttt{\{claim\_text\}}\par
\par\smallskip
\noindent Return exactly one label object per claim id.\par
\end{tcolorbox}\end{nolinenumbers}

\subsection{Follow-Through Verification}\label{app:python-agent-verification}

\begin{nolinenumbers}\begin{tcolorbox}[vftprompt, colframe=black, colback=white, colbacktitle=white, coltitle=black, title={Follow-Through Verification Prompt}, title after break={Follow-Through Verification Prompt (continued)}]
\textbf{\textsc{System message}}\par\smallskip
\noindent You are a securities analyst checking FOLLOW-THROUGH: did a company realize the commitment stated in its IPO prospectus (S-1 or F-1)? Assess the original claim against its scope, any stated deadline, and the dates and periods covered by the supplied evidence. You are given the claim, candidate SEC passages with source metadata, and optional annual SEC XBRL trend context or cached Metric Memory. Each passage is labeled [P\#\#] with its source. Using ONLY this SEC evidence, assign exactly ONE outcome:\par
\begin{VFTPromptList}
\item delivered: evidence establishes realization consistent with the claim's scope and applicable deadline.
\item partially\_delivered: evidence establishes some, but not all, of the claimed realization by the date covered by the evidence.
\item not\_delivered: sufficient evidence establishes non-realization by the applicable deadline or assessment date. Missing expected evidence or a discussion of the topic alone is insufficient.
\item reversed: evidence explicitly establishes that the firm abandoned or contradicted the commitment.
\item unverifiable (Indeterminate): evidence is insufficient, the relevant period has not yet elapsed, or the claim lacks an assessable condition. Record which reason applies. Do not infer Not Delivered from silence, an unexpired deadline, or retrieval failure.
\end{VFTPromptList}
\noindent Check source dates and reporting periods against the claim and supplied assessment cutoff. Do not invent a missing deadline, date, or evidence window. If missing temporal information prevents assessment, return unverifiable and explain the gap. Do not treat evidence that repeats a commitment as evidence of its realization. Use XBRL trends and Metric Memory for longitudinal context, but outcome support must come from cited passages.\par
\noindent Cite the passage id(s) you relied on and copy a VERBATIM quote from each cited passage --- exact characters, no paraphrasing. For unverifiable, cite relevant passages if they explain the unresolved status; return an empty evidence list if no passage bears on the claim. Never fabricate a quotation to fill the schema. Give a confidence in [0,1] and a concise rationale tied to the evidence and applicable time period.\par
\tcblower\textbf{\textsc{User message}}\par\smallskip
\par\noindent\textbf{Claim (from the IPO prospectus):}\par\nobreak
\noindent "\texttt{\{unit\_text\}}"\par
\noindent Claim date: \texttt{\{claim\_date\}}\par
\noindent Stated deadline, if any: \texttt{\{claim\_deadline\}}\par
\noindent Assessment date and evidence cutoff: \texttt{\{assessment\_cutoff\}}\par
\noindent What to look for: \texttt{\{expected\_evidence\}}\par
\par\smallskip
\par\noindent\textbf{Annual SEC XBRL trend context:}\par\nobreak
\noindent \texttt{\{xbrl\_context\}}\par
\par\smallskip
\par\noindent\textbf{Cached Metric Memory:}\par\nobreak
\noindent \texttt{\{metric\_context\}}\par
\par\smallskip
\par\noindent\textbf{Candidate passages from later filings:}\par\nobreak
\noindent \texttt{\{labeled\_passages\}}\par
\par\smallskip
\noindent Record your verdict.\par
\end{tcolorbox}\end{nolinenumbers}
{\small\noindent \textbf{Web-augmented variant.} The system replaces \texttt{Using ONLY this SEC evidence} with \texttt{Using ONLY the evidence provided below}, then appends the following instruction. All other system text is unchanged.\par}
\begin{nolinenumbers}\begin{tcolorbox}[vftprompt, colframe=black, colback=white, colbacktitle=white, coltitle=black, title={Web Evidence Extension}, title after break={Web Evidence Extension (continued)}]
\textbf{\textsc{System message}}\par\smallskip
\noindent Some passages are labeled [W\#\#]: these are OPEN-WEB sources (news, press releases, analyses), not SEC filings. You may use them to corroborate or contradict a proposition and may cite them as proofs, but treat SEC filings ([P\#\#]) as the primary, authoritative source and prefer them when they suffice. Web quotes must still be VERBATIM from the cited [W\#\#] passage.\par
\end{tcolorbox}\end{nolinenumbers}
{\small\noindent The user heading becomes \texttt{Candidate passages (SEC filings [P\#\#] + open web [W\#\#]):}. OpenSERP, You.com, SerpAPI, and provider search affect retrieval, not this template. Filings-only evaluations must disable the available web fallback.\par}

\subsection{Conversational Controller}\label{fig:agent_prompt}

\begin{nolinenumbers}\begin{tcolorbox}[vftprompt, colframe=black, colback=white, colbacktitle=white, coltitle=black, title={Conversational Controller Prompt}, title after break={Conversational Controller Prompt (continued)}]
\textbf{\textsc{System message}}\par\smallskip
\noindent You are VFT --- "Vision Follow Through" --- a finance-native terminal agent. Your specialty: deciding whether a company actually delivered on the promises it made in its IPO prospectus (SEC Form S-1/F-1).\par
\par\smallskip
\par\noindent\textbf{THE VFT PIPELINE}\par\nobreak
\par\noindent\textbf{The harness extracts original claims from an S-1:}\par\nobreak
\begin{VFTPromptList}
\item CLAIMS: discrete assertions the issuer makes --- about its market (MO), product/technology (PT), financials/growth (FG), organization/culture (OC), or mission (SM). Each carries a composite label CAT-VER[-NEG], where VER is how checkable it is: DV (Directly Verifiable --- a later 10-K can settle it), IV (Indirectly Verifiable), or A (Aspirational). -NEG marks a self-limiting claim.
\end{VFTPromptList}
\par\smallskip
\par\noindent\textbf{HOW TO WORK}\par\nobreak
\begin{VFTPromptList}
\item When the user names a company, call dissect\_company. It is CACHE-FIRST: if VFT already analyzed it, results load instantly --- so always reach for it rather than guessing. Use list\_companies to see what's already cached.
\item Use get\_claims / company\_overview to inspect the analysis.
\item Use label\_claims when the user asks for Task 2, independent category assignment, or independent verifiability assignment. It classifies already-extracted claim spans separately and caches labels without overwriting extraction outputs.
\item VERIFY follow-through with verify\_company. Use verification\_mode=direct\_claims to assess each original extracted claim. Inspect the actual retrieved forms, source dates, and observation period; do not assume that default retrieval enforces IPO-relative horizons or a historical web cutoff. Distinguish delivered (full realization), partially\_delivered (some realization), not\_delivered (sufficient evidence of non-realization by the applicable deadline or assessment date), reversed (explicit abandonment or contradiction), and unverifiable (the paper's Indeterminate outcome). Missing evidence alone is not non-delivery. Keep execution failures separate from judgments. Cite verbatim evidence for supported outcomes and explain evidence gaps for Indeterminate judgments. After verifying, call get\_verdicts with the same verification\_mode to surface the evidence.
\item Summarize the VFT score as a weighted mean over assessable claims: Delivered +1, Partially Delivered +0.5, Not Delivered 0, Reversed -0.5, with DV/IV/A weights 1.0/0.6/0.3. Exclude Indeterminate judgments and report assessable and total counts, coverage, and the per-category profile. If no claims are assessable, report the score as undefined. These weights are modeling choices, not a calibrated measure of corporate credibility.
\item Use evaluate\_company when the user gives a filled human-annotated claim file and asks for benchmark metrics. It compares agent artifacts to human-annotated claims for extraction, category, verifiability, outcome, and VFT deltas without rerunning the model.
\item Use metric\_memory when claims hinge on recurring operating KPIs like ARR, NRR/DBNRR, customers over \$100k, gross margin, contribution margin, cohorts, RPO, or billings. Treat it as the company's normalized KPI memory across filings.
\item sec\_financials and list\_filings pull the raw reported numbers and filing list directly. Separate fact from estimate; never invent numbers --- let the proofs do the talking.
\end{VFTPromptList}
\par\smallskip
\begin{VFTPromptList}
\item Use longitudinal\_followthrough for year-over-year, multi-year, trend, growth, profitability, retention, expansion, or product-adoption questions. Start with a focused query, inspect the annual XBRL trend line, then refine with another query if the filing language is still thin.
\end{VFTPromptList}
\par\smallskip
\par\noindent\textbf{STYLE}\par\nobreak
\begin{VFTPromptList}
\item You run in a terminal and your tools already render rich, beautiful cards, tables, and dashboards to the user. So DO NOT repeat long lists back in prose --- the user can see them. Add a tight, insightful summary: the 2-4 things that matter, a number that stands out, a follow-through judgment.
\item Format with Markdown --- the terminal renders it richly. Use short **bold** section headings, "-" bullets and numbered lists, and "---" rules to structure a verdict. When you summarize follow-through, a heading that ends in \ding{51} (delivered), \ding{55} (broken), or $\triangle$ (partial) gets auto color-coded --- use them.
\item Be concise and well-structured. You give analysis and education, not personalized investment advice.
\end{VFTPromptList}
\end{tcolorbox}\end{nolinenumbers}

\endgroup

\section{\vftagent{} Implementation Details}
\label{app:pipeline-details}

This appendix specifies the Python components behind
Section~\ref{sec:pipeline}. Runtime defaults describe the implementation;
the configuration saved with each experiment determines its protocol.
Appendix~\ref{app:prompts} provides the prompts.

\subsection{Execution and Stage Outputs}
Table~\ref{tab:agent-execution} follows a claim from its source filing to
its assessment. The interactive controller selects tools over at most
12 model turns, executing requested tools sequentially within each turn.
The benchmark invokes the components directly in a fixed sequence,
using either extracted claims or supplied human-annotated claims as input.

\begin{table*}[t]
\centering\small
\renewcommand{\arraystretch}{1.16}
\begin{tabularx}{\textwidth}{p{2.0cm}X p{4.1cm}}
\toprule
\textbf{Stage} & \textbf{Operation} & \textbf{Inspectable output} \\
\midrule
\textbf{1. Extract} &
Read the IPO filing; select numbered source units and reconstruct claim spans. &
\path{claims.jsonl}, source and unit identifiers \\
\textbf{2. Label} &
Assign a subject category and verifiability tier to each existing claim. &
\path{claim-labels.json}, confidence and rationale \\
\textbf{3. Retrieve} &
Collect SEC filings and, when enabled, You.com web results; rank passages. &
Passage IDs, document metadata and source URLs \\
\textbf{4. Verify} &
Assess each original claim against the selected evidence and optional metric context. &
Outcome, rationale, confidence and quotations \\
\textbf{5. Report} &
Save assessments; summarize assessable claims, coverage and issuer scores. &
\path{verdicts-direct-claims.json}, scorecard and coverage summary \\
\bottomrule
\end{tabularx}
\caption{Claim-based execution flow shared by interactive tools and benchmark
tasks. Flyte observes interactive requests or orchestrates cached
document-level benchmarks.}
\label{tab:agent-execution}
\end{table*}

\paragraph{Extraction and labeling.}
The Python segmenter splits filing text into numbered, sentence-like
units and preserves their character offsets. Long inputs are batched;
length-truncated model responses can trigger smaller batches. The model
selects unit identifiers and proposes categories; claim text is then
reconstructed from the source. Separate labeling assigns a category and
DV, IV, or A tier, with confidence, negation, and rationale, without
changing claim text. The interactive limit is 60 claims per labeling
call; the benchmark batch-size default is 10. Verification assesses these
claims directly.

\subsection{Evidence Acquisition and Ranking}
\paragraph{Source collection.}
SEC EDGAR supplies filing metadata, documents, and XBRL financial
data.\footnote{\url{https://www.sec.gov/search-filings/edgar-application-programming-interfaces}}
The default benchmark requests Forms 10-K and 20-F with selector values
\texttt{years=[1,3]}. For each form, these select filings closest to the
earliest returned filing and two years after it; they do not establish
one- and three-year horizons from the IPO. An alternative interactive
spacing mode can also include 10-Q and 6-K filings. Downloaded text is
segmented into approximately 700-character passages, splitting long
paragraphs at sentence boundaries. Unusable filings are recorded.
XBRL trends may supply additional context, but quotations must come from
the retrieved passages.

\paragraph{Optional web evidence.}
You.com is the default web-search backend.\footnote{\url{https://you.com/docs/api-reference/search/v1-search}}
Queries combine the company name and claim text. The search excludes
\texttt{sec.gov}, deduplicates URLs, and attempts to download the first
three non-SEC results. Their page text is segmented and added to the SEC
passage pool. Benchmark SEC and web switches
are independent; \texttt{--no-web} disables web retrieval. Interactive
verification can also request web evidence when SEC documents are absent.

\paragraph{Hybrid passage ranking.}
For a claim query $q$, BM25 \citep{10.1561/1500000019} ranks passages by
lexical relevance with $k_1=1.5$ and $b=0.75$. A second ranking uses cosine
similarity between normalized sentence embeddings
\citep{reimers-gurevych-2019-sentence}. The default checkpoint is
\texttt{all-mpnet-base-v2}, a sentence-transformer model built on MPNet
\citep{song2020mpnet}.\footnote{Checkpoint and model card: \url{https://huggingface.co/sentence-transformers/all-mpnet-base-v2}}
RRF \citep{cormack2009rrf} combines the two rankings:
\begin{equation}
 s(p,q)=\sum_{r\in\{\mathrm{BM25},\mathrm{dense}\}}
 \frac{1}{60+\operatorname{rank}_{r}(p,q)}.
\end{equation}
Ranks start at one; a passage absent from a ranking contributes zero.
The top six passages are retained for SEC-only verification, or the top
eight across the combined SEC and web pool. This is a shared budget,
not an additional allocation to web evidence. Passage identifiers link each excerpt to its filing metadata
or web URL. 

\begin{table*}[t]
\centering\small
\renewcommand{\arraystretch}{1.12}
\begin{tabularx}{\textwidth}{p{0.65cm}p{2.6cm}XX}
\toprule
\textbf{Task} & \textbf{Input} & \textbf{Required output} & \textbf{Evaluation} \\
\midrule
1 & IPO filing & Claim spans and source locations (whole-unit selections in the baseline) &
Exact-match and overlap-based span precision, recall, and F$_1$ \\
2 & Claim and IPO context & Primary category and verifiability tier &
Separate macro-F$_1$ and per-class metrics on available human-annotated labels \\
3 & Claim, issuer, horizon, and admissible evidence collection &
Source documents, quoted passages, justification, and claim-level outcome
or Indeterminate status & Document/passage Recall@$k$, grounding,
outcome metrics, end-to-end accuracy, and coverage \\
\bottomrule
\end{tabularx}
\caption{VFT task definitions. Claim-level outputs define the benchmark;
retrieval budgets and orchestration settings are implementation choices.
The evaluation column specifies the evaluation scope; not every listed
metric is reported in Section~\ref{sec:results}.}
\label{tab:vft-tasks}
\end{table*}

\subsection{Assessment and Validation}
The judge receives the original claim, retrieved passages, and available
XBRL context. It returns structured JSON containing an outcome, confidence,
rationale, and evidence entries with passage identifiers and quotations.
The implementation outcomes are \texttt{delivered},
\texttt{partially\_delivered}, \texttt{not\_delivered},
\texttt{reversed}, and \texttt{unverifiable}.
The paper's corresponding outcome labels are Delivered, Partially
Delivered, Not Delivered, Reversed, and Indeterminate. This label mapping
does not make an execution failure a completed Indeterminate judgment.

Interactive and benchmark validation differ. The interactive tool drops
unknown passage references and may replace an unmatched quotation with
a source sentence. The benchmark judge instead validates the response
schema, restricts references to the supplied passages, and requires each
quotation to be a source substring after whitespace normalization. It
makes up to three attempts before recording invalid output. A claim
without passages receives a separate no-evidence status without a judge
call. These checks establish traceability, not semantic entailment;
grounding evaluation must still assess whether the evidence supports
the outcome.

\subsection{Flyte Execution and Reuse}
Flyte provides task orchestration.\footnote{\url{https://www.union.ai/docs/v2/flyte/user-guide/tasks/task-configuration/caching/}}
The local interactive observer records tool activity, timing, status,
and available usage without altering controller execution or enabling
Flyte caching or retries. The durable benchmark instead runs cached
document tasks with two retries and aggregates their outputs. Run
artifacts retain data, prompt, configuration, and implementation
fingerprints alongside predictions, evidence, and metrics.

Compatible saved outputs support incremental extraction--labeling,
extraction--labeling--verification, and human-annotated claim verification. Evidence
reuse requires matching document, normalized claim text, issuer metadata,
retrieval settings, and evidence-collection code. Judge-response reuse
additionally requires matching model, prompt, and generation settings;
metrics are recomputed against the current human-annotated labels. This allows
different judges to use the same evidence snapshot. Interactive verdict
reuse is separate and requires an explicit refresh after configuration
changes. 

\subsection{Verification Coverage and Scoring}
\label{app:verification-coverage}

Table~\ref{tab:verification-coverage} reports judgment coverage and
claim-level verification accuracy across models. We report accuracy over
all 1,000 claims and separately over claims receiving valid judgments.

\begin{table*}[t]
\centering
\small
\setlength{\tabcolsep}{5pt}
\renewcommand{\arraystretch}{1.15}

\begin{tabularx}{\textwidth}{
    @{}>{\raggedright\arraybackslash}Xrrrrr@{}
}
\toprule
\textbf{Model}
& \shortstack{\textbf{Coverage}\\\textbf{(\%)}}
& \shortstack{\textbf{Invalid}\\\textbf{responses ($n$)}}
& \shortstack{\textbf{Accuracy}\\\textbf{all 1,000 (\%)}}
& \shortstack{\textbf{Accuracy}\\\textbf{judged only (\%)}}
& \shortstack{\textbf{Macro-F$_1$}\\\textbf{all claims (\%)}} \\
\midrule
Claude Haiku 4.5
& 67.0 & 195 & 35.1 & 52.4 & 32.5 \\
Gemma 4 31B
& 86.1 & 4   & 44.1 & 51.2 & 36.2 \\
Llama 3.1 8B
& 62.0 & 245 & 18.1 & 29.2 & 12.1 \\
Llama 3.1 70B
& 83.9 & 26  & 41.8 & 49.8 & 35.3 \\
GPT-oss 20B
& 83.6 & 29  & 36.9 & 44.1 & 31.4 \\
GPT-oss 120B
& 84.3 & 22  & 38.2 & 45.3 & 32.2 \\
GPT-6 Luna
& 86.4 & 1   & 38.4 & 44.4 & 29.3 \\
Nemotron 3.5 Super VL
& 85.4 & 11  & 40.4 & 47.3 & 29.9 \\
\bottomrule
\end{tabularx}

\caption{
Verification coverage and scoring for
Table~\ref{tab:verification-results}.
All models evaluated 1,000 claims using identical SEC and
You.com evidence.
Coverage is the percentage of valid judgments.
Unjudged cases include claims which do not contain follow-through evidence in subsequent documents and invalid
responses; Indeterminate is a valid judgment.
All-claims accuracy and macro-F$_1$ penalize unjudged cases,
while judged-only accuracy excludes them.
Open-weight models use Together; Claude Haiku 4.5 uses Anthropic;
GPT-6 Luna uses OpenAI with \texttt{reasoning\_effort=none}.
Nemotron and Claude Haiku 4.5 use disabled reasoning.
All runs have an 8,192-token output limit.
}
\label{tab:verification-coverage}
\end{table*}

\section{Failure Examples}
\label{app:diagnostic-examples}

Below are the failure categories. Table~\ref{error-examples} provides examples for failure categories.

\paragraph{Inconsistent claim criteria.}
The extraction instructions and human-annotated claim construction admit different kinds
of statements. An output can therefore follow the instructions while
falling outside the benchmark's retained claims.

\paragraph{Omitted claims.}
The model fails to extract any part of a commitment identified in the
human-annotated claims.

\paragraph{Incomplete claims.}
An extraction captures only part of a claim, omitting content needed
to interpret what is promised.

\paragraph{Merged claims.}
A single extracted claim combines distinct commitments that could each
require a separate follow-through assessment. A single outcome can then
conceal that one commitment was met while another remains unresolved.

\paragraph{Discontinuous claims.}
An output joins nonadjacent source units into a claim that cannot be
mapped to one contiguous passage in the filing. This breaks the link
between the proposition being assessed and the company's original wording.

\paragraph{Granularity mismatch.}
Predicted and human-annotated claim spans segment the same source content differently.
Under one-to-one matching, multiple qualifying predictions can compete
for a single human-annotated claim span, leaving some unmatched despite covering relevant
text.

\paragraph{Category misclassification.}
A claim receives a category that differs from its human-annotated label, reflecting
a disagreement about the subject or objective represented by the claim.

\paragraph{Truncated generation.}
Generation reaches its output limit before completing the response.

\begin{table*}[t]
\centering\small
\setlength{\tabcolsep}{4pt}
\renewcommand{\arraystretch}{1.08}
\begin{tabularx}{\textwidth}{@{}>{\raggedright\arraybackslash}p{0.13\textwidth}>{\raggedright\arraybackslash}p{0.09\textwidth}>{\raggedright\arraybackslash}p{0.42\textwidth}>{\raggedright\arraybackslash}X@{}}
\toprule
\textbf{Error category} & \textbf{Model} & \textbf{Human-annotated claims and model behavior} & \textbf{Error explanation} \\
\midrule
\textbf{Inconsistent claim criteria} &
Gemma 4 31B &
\textbf{Human-annotated claim:} No intersecting span.\newline
\textbf{Model-extracted claim:} ``in the quarter ended December 31, 2018, 50\% of first-time [\ldots] consumers were new to our platform.'' &
The guide permits some historical assertions, so this unmatched statement is not necessarily invalid. The example exposes a mismatch between instructions and benchmark coverage. \\
\addlinespace[5pt]
\textbf{Omitted claims} &
GPT-6 Luna &
\textbf{Human-annotated claim:} ``The goal of our board of directors is [\ldots] senior leadership development and succession.''\newline
\textbf{Model-extracted claim:} No located prediction intersects this span. &
The model omits a commitment to leadership succession that the guide includes under Organizational/Culture. \\
\addlinespace[5pt]
\textbf{Incomplete claims} &
GPT-6 Luna &
\textbf{Human-annotated claim:} ``[\ldots] Four-Wall Margins of 35\% and Average Sales Per Square Foot of \$2,900.''\newline
\textbf{Model-extracted claim:} ``We expect Average Sales Per Square Foot may fluctuate due to factors including [\ldots]'' (token-F$_1=0.62$). &
The output drops both numerical targets but still meets the overlap threshold, omitting the measurable commitment. \\
\addlinespace[5pt]
\textbf{Merged claims} &
GPT-6 Luna &
\textbf{Human-annotated claim:} ``We expect intense competition to continue [\ldots]''\newline
\textbf{Model-extracted claim:} the same text, plus ``We expect advertising spend to continue to shift from traditional linear TV to CTV [\ldots]'' (token-F$_1=0.69$). &
The extraction merges forecasts about competition and advertising shifts that require separate assessments. \\
\addlinespace[5pt]
\textbf{Discontinuous claims} &
GPT-6 Luna &
\textbf{Human-annotated claim:} No intersecting span.\newline
\textbf{Model-extracted claim:} ``Our Compensation Committee will executive officers going forward.'' (units U6758 and U6765) &
Claims should come from one contiguous passage. Joining nonadjacent units skips six units, deletes the action, and produces wording absent from the source. \\
\addlinespace[5pt]
\textbf{Granularity mismatch} &
Gemma 4 31B &
\textbf{Human-annotated claim:} One span covering sales-force investment and increasing sales/marketing expenses.\newline
\textbf{Model-extracted claim:} Two claims, ``[\ldots] invest in our sales force [\ldots]'' and ``expect our sales and marketing expenses to increase [\ldots]'' &
Both fragments meet the overlap threshold, but only one can match the reference span. The outputs are distinct, not duplicates. \\
\addlinespace[5pt]
\textbf{Category misclassification} &
GPT-6 Luna &
\textbf{Matched claim:} ``[\ldots] expand the applicability of our platform to creators with new solutions [\ldots] such as assisted artistry workflows, higher performance rendering capabilities [\ldots]''\newline
\textbf{Human-annotated category:} Market Opportunity.\newline
\textbf{Model category:} Product/Technology. &
The model emphasizes the product mechanism; the human annotation emphasizes the market expansion objective. Both appear in the claim. \\
\addlinespace[5pt]
\textbf{Truncated generation} &
Nemotron 3 Nano 4B &
\textbf{Human-annotated claim:} Not applicable \newline
\textbf{Model response:} ends mid-string with \texttt{[\ldots] "U1469", "U1470", "U} (\texttt{finish\_reason=length}; 8,192 completion tokens) &
The response should be complete, schema-valid JSON, but the token limit cuts it off mid-string. This failed attempt does not establish failure of the final document evaluation. \\
\bottomrule
\end{tabularx}
\caption{Error analysis examples from the final 5,066-claim evaluation.
\textbf{Human-annotated claim}: the retained human span
(``No intersecting span'' reflects benchmark matching, not a semantic
judgment; ``Not applicable'' where no claim was produced).
\textbf{Model-extracted claim}: the model extraction that overlaps the
human-annotated claim, or the unmatched extraction when none overlaps.
\textbf{Model response}: raw output when no claim was produced.
\textbf{Matched claim}: a retained human-annotated span matched by the model,
with its \textbf{Human-annotated category} and \textbf{Model category}.
Token-F$_1$ measures overlap with the matched human-annotated claim;
U-numbers are source-unit identifiers; [\ldots] marks omitted text.}
\label{error-examples}
\end{table*}

\end{document}